# Motion Planning by Reinforcement Learning for an Unmanned Aerial Vehicle in Virtual Open Space with Static Obstacles

Sanghyun Kim[1], Jongmin Park[1], Jae-Kwan Yun[2], and Jiwon Seo[1*]

[1] School of Integrated Technology, Yonsei University,
Incheon, 21983, Korea (sanghyun.kim, jm97, jiwon.seo@yonsei.ac.kr)
[2] Electronics and Telecommunications Research Institute,
Daejeon, 34129, Korea (jkyun@etri.re.kr)
* Corresponding author

**Abstract**: In this study, we applied reinforcement learning based on the proximal policy optimization algorithm to perform motion planning for an unmanned aerial vehicle (UAV) in an open space with static obstacles. The application of reinforcement learning through a real UAV has several limitations such as time and cost; thus, we used the Gazebo simulator to train a virtual quadrotor UAV in a virtual environment. As the reinforcement learning progressed, the mean reward and goal rate of the model were increased. Furthermore, the test of the trained model shows that the UAV reaches the goal with an 81% goal rate using the simple reward function suggested in this work.

## 1. INTRODUCTION

In recent years, unmanned aerial vehicles (UAVs) have become widely used in civil applications such as transportation, surveillance systems, and environmental protection [1-5]. The navigation of various vehicles typically relies on Global Navigation Satellite Systems (GNSSs) [6-9] and other sensors [10-15]. However, the motion planning for UAVs remains challenging because the external environment is unpredictable [16]. Reinforcement learning can be applied to perform motion planning, and allows selection of the optimal action according to the current state [17-20].

In this paper, we propose a method of motion planning for a UAV by reinforcement learning in an open space with static obstacles. Among the various reinforcement learning algorithms, including Deep-Q-Network, Q-Learning, and proximal policy optimization (PPO) [21-23], we used the PPO algorithm as it is most suitable for continuous control tasks [24].

Reinforcement learning typically requires a large amount of data; however, in the case of UAVs, there are practical limitations such as time and cost to obtaining a large training dataset in the real environment [25, 26]. Therefore, by using the Gazebo simulator [27], we implemented a virtual environment that follows the same physical laws as the real environment and performed reinforcement learning in the virtual environment. To demonstrate the performance of the proposed method, the obtained learning model was tested in the simulation environment.

## 2. UAV SIMULATION ENVIRONMENT

The UAV simulation environment of this work comprises three software: Gazebo, Robot Operating System (ROS), and OpenAI Gym. Gazebo is a 3D dynamic simulator that helps simulating robots in the virtual environments. Fig. 1 shows the models that are rendered by Gazebo, which represents an open space with static obstacles and a quadrotor.

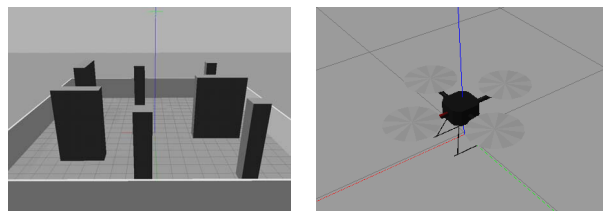

Fig. 1 Open space with static obstacles (left) and the quadrotor (right) implemented for this study.

ROS is an open source meta operating system for robots, which provides services such as hardware abstraction, low-level device control, and message-passing between processes [28]. In our simulation environment, ROS controls the movement of the quadrotor in the Gazebo environment and enables information transfer between Gazebo and OpenAI Gym through message communication. OpenAI Gym is a toolkit that includes various simulation environments to test reinforcement learning algorithms [29]. Reinforcement learning algorithms can also be developed in OpenAI Gym environment, and it is compatible with reinforcement learning libraries such as Tensorflow and RLlib. We used RLlib [30] to perform reinforcement learning in the Gym environment.

These three software have the following relationship in the simulation environment and perform the following simulation process: The virtual quadrotor and open space environment are rendered in Gazebo, and the virtual sensor measurements of the quadrotor obtained by Gazebo are transferred to the OpenAI Gym environment through ROS. Then, the reinforcement learning algorithm is executed in the Gym environment based on the state of the quadrotor, and the appropriate action according to the state is transmitted to the ROS to control the quadrotor movement.

## 3. LEARNING ENVIRONMENT

The reinforcement learning algorithm used in this study is PPO, implemented through a function named PPOTrainer provided by RLlib. In reinforcement learning, the agent judges the current state based on the state space, and then selects and performs one action that can maximize the reward from various actions defined in the action space. The time unit corresponding to the above process is called a step, and several steps gather to form an episode. During one episode, the quadrotor performs one task, which is to reach the goal while avoiding obstacles. Multiple episodes are gathered to form one iteration, and when one iteration is complete, the model parameters related to reinforcement learning are updated. The goal rate was calculated for each iteration, and if the goal rate exceeds 80% in five consecutive iterations, it is set to end the learning.

### 3.1 State Space

The state space is defined as shown in Table 1.

Table 1 State space of the reinforcement learning model.

|  | State Space |
|---|---|
| Depth | Depth camera 2D image (9×12) |
| Heading | Angle difference between the heading of the quadrotor and the goal [rad] |
| Distance | Euclidean distance between the quadrotor and the goal [m] |

### 3.2 Action Space

The action space is defined as shown in Table 2.

Table 2 Action space of the reinforcement learning model.

|  | Action Space |
|---|---|
| Forward Velocity | Forward velocity among 0, +1, +2 [m/s] |
| Yaw Rate | Yaw rate among -pi/2, -pi/4, 0, +pi/4, +pi/2 [rad/s] |

### 3.3 Learning Parameters

The default value of the PPOTrainer function was used as the learning parameter setting, except the batch size was changed and 'batch_mode' was modified to 'complete_episodes'. The batch size was set to 10,000 steps.

### 3.4 Rewards

The reward shaping process focused on the following three objectives: reaching the goal, avoiding collision with obstacles, taking as little time as possible to perform the task.

To allow the quadrotor to reach the goal, at every step, the quadrotor receives a positive reward for every step closer to the goal, and a large positive reward when the quadrotor reaches the goal. Meanwhile, there is a large negative reward when the quadrotor collides with the obstacles. In addition, the quadrotor is given a small constant negative reward for each step, to ensure the task is performed in as short a time as possible. At every step, the quadrotor receives the sum of all these rewards. The reward at time $t$ can be expressed as follows:

$$r_t = scaling\ factor(20) \times (d_t(q,g) - d_{t-1}(q,g)) \\ + 2000 \times \mathbb{1}[goal\_reached] \\ - 1000 \times \mathbb{1}[collision] - 1 \qquad (1)$$

where $q$ denotes the quadrotor, $g$ denotes the goal, $d_t(\cdot,\cdot)$ is the Euclidean distance between two parameters at time $t$, and $\mathbb{1}$ is the indicator function whose value is 1 when its argument is true.

## 4. SIMULATION RESULTS

Fig. 2 shows the coordinate sets as rectangles, which represent the possible sets of the start and goal points of the quadrotor. The coordinate sets represented by the red and blue rectangles were configured for training and testing, respectively. When the episode is triggered, two random coordinates in the set are selected as the start and goal points.

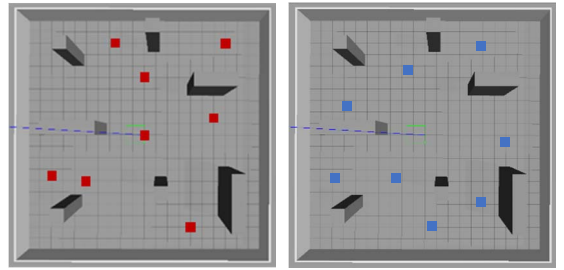

Fig. 2 Coordinate sets of the start and goal points of the quadrotor: red (left) and blue (right) rectangles represent the coordinate sets for training and testing, respectively.

Figs. 3 and 4 illustrate the training results, showing the mean reward and the goal rate per iteration, respectively. Training took a total of 54 iterations, and as seen in these figures, both the mean reward and goal rate increase with the iterations.

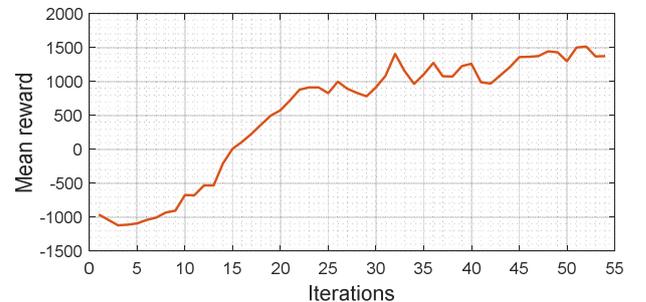

Fig. 3 Mean reward with respect to the number of iterations.

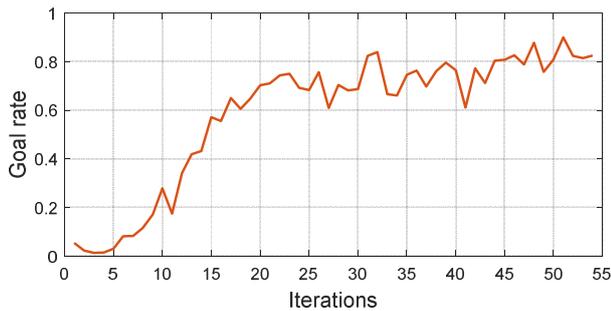

Fig. 4 Goal rate with respect to the number of iterations.

When the trained model was tested, the quadrotor reached the goal without collision 81 times out of 100 episodes, resulting in a goal rate of 81%.

## 5. CONCLUSION

In this study, we applied reinforcement learning based on the PPO algorithm in a virtual environment to perform motion planning for a UAV. The virtual open space environment with static obstacles was implemented using the Gazebo simulator. The trained model demonstrated the motion planning capability with the goal rate of 81%.

## ACKNOWLEDGEMENT

This work was supported by Electronics and Telecommunications Research Institute (ETRI) grant funded by the Korean government [20ZR1100, Core Technologies of Distributed Intelligence Things for Solving Industry and Society Problems].